\documentclass[Journal,letterpaper,SingleSpace,InsideFigs,NoLineNumbers,NoLists]{ascelike-new}
\usepackage[utf8]{inputenc}
\usepackage[T1]{fontenc}
\usepackage{lmodern}
\usepackage{graphicx}
\usepackage[style=base,figurename=Fig.,labelfont=bf,labelsep=period]{caption}
\usepackage{amsmath}
\usepackage{booktabs}
\usepackage{enumitem}
\usepackage{newtxtext,newtxmath}
\usepackage[colorlinks=true,citecolor=black,linkcolor=black,urlcolor=black]{hyperref}
\graphicspath{{figs/}}
\NameTag{}

\begin{document}

\title{Silence Is Not Assurance: When Production Lane Keeping Assist Does Not Warn, and What It Costs}

\author[1]{Tianjia Yang}
\author[1]{Yuxin Ding}
\author[2]{Ke Li}
\author[2]{Kaidi Liang}
\author[2]{Ruwen Qin}
\author[1,*]{Xianbiao Hu}

\affil[1]{Department of Civil and Environmental Engineering, The Pennsylvania State University, University Park, PA 16802, USA}
\affil[2]{Department of Civil Engineering, Stony Brook University, Stony Brook, NY 11794, USA}
\affil[*]{Corresponding author. Email: xbhu@psu.edu}

\maketitle

\begin{abstract}
Production lane keeping assist (LKA) systems tell the driver whether they detect the lane markings, but whether this displayed status reflects what the system perceives, and what happens when it does not, has not been examined in real-world driving. This study introduces CANDID-LKA, a CAN-based Diagnosis of the Indicated Detection status in LKA. It evaluates a proprietary driver-facing status using three signals: the displayed status and the driver's steering torque, both recorded from the vehicle's controller area network (CAN) bus, and the lane perception confidence of a third-party vision system observing the same road. The framework examines how the status responds to driving conditions and estimates how completely it reports degraded lane perception by comparing it with the other two signals in a latent-class model. We apply it to 39{,}800 seconds of real-world driving from 13 vehicle models of three manufacturers. The results show that the status does not respond consistently to the conditions that make markings harder to see, and that the three makes differ widely in how often they report \emph{not detected}. The latent-class model shows that the displayed status is incomplete: it reports only 47\% of degraded seconds as \emph{not detected}, compared with 80\% for the third-party low perception confidence, and in half of the low-confidence episodes it never reports \emph{not detected}. In the silent seconds in which the status reads \emph{detected} while the perception confidence is low and the driver is not in control, the standard deviation of lateral position over the next 3~s is about 60\% larger than when the status and the perception confidence agree. A \emph{detected} status thus does not guarantee reliable lane perception, and silence with the driver not in control is the case that most needs attention.

\end{abstract}

\section{Introduction}\label{sec:intro}

Lane keeping assist (LKA) has become a standard function on most new passenger vehicles in the United States. Adoption rose from 1.4 percent of model year 2015 vehicles to 86.3 percent of model year 2023 vehicles \cite{mitre2024parts}, and the function is associated with measurable reductions in lane-departure crashes \cite{cicchino2018lane,dean2022estimating}. Those benefits are not uniform. LKA relies on camera-based perception of the road ahead, so its performance can degrade where markings are worn, contrast is low, or weather obscures the lane \cite{re2021testing,neumeister2019adverse}, and the benefit realized also depends on how drivers use the system \cite{cicchino2025convenience}. A driver cannot directly observe the system's internal perception. The lane-detection status on the dashboard is the vehicle's primary explicit report of whether it currently recognizes the lane. This paper evaluates the trustworthiness of that report, the extent to which a driver can rely on it as an indication of the current condition of lane perception, and what it costs the driver when the report stays silent.

The displayed status can diverge from the underlying perception condition in two ways, with different consequences for the driver. In the first case, the status reports the lane as \emph{not detected} while perception is adequate. The cost is a nuisance alert, and if such alerts repeat, the cue loses credibility and drivers may discount it or switch the function off \cite{wilson2020trust,wischnewski2023trust,mueller2024habits}. In the second case, the status keeps showing \emph{detected} after perception has degraded. The display continues to indicate normal operation, so the driver may keep relying on the system and respond too late to a developing lane departure. The latter is the more damaging of the two, because the driver is given no signal at the moment a lane departure is most likely.

Prior research has characterized when LKA perception succeeds and fails, but the trustworthiness of the displayed status has not itself been evaluated. Perception performance has been studied through controlled and instrumented-road testing, simulation of rare and severe conditions, and large-scale naturalistic evaluation of production vehicles \cite{re2021testing,yang2025simlkas,alhuraish2025empirical}. A related line of work examines whether automated driving systems can recognize their own perception limits through runtime monitoring, out-of-distribution detection, and uncertainty estimation \cite{stocco2020misbehaviour,henriksson2023ood,feng2022probabilistic}. In parallel, human-factors research investigates how drivers calibrate their reliance on partial automation from system feedback and documents the mode confusion and miscalibrated trust that can arise when that feedback is misleading \cite{wilson2020trust,wischnewski2023trust,mueller2024habits}. What is missing is an evaluation of the signal that connects them, the status the vehicle actually shows the driver.

Evaluating the trustworthiness of the displayed status faces two challenges. The first is access. The displayed status is a report produced by the system, not a direct view of what the system perceives. Production perception systems are proprietary, so what the system internally perceives cannot be observed, and only the detection state it reports on the vehicle bus is available. Any evaluation must therefore work from the report and from signals recorded alongside it, not from the perception itself. The second is the absence of a reference. Whether the lane markings are perceivable at a given moment is not a deterministic outcome, and no single agreed standard exists against which the displayed status could be checked directly. Marking perceivability depends on wear, contrast, lighting, weather, and the camera, and it varies continuously, so no ground-truth label of perceivable or not perceivable exists for each second of driving.

This paper addresses both challenges with CANDID-LKA, a CAN-based Diagnosis of the Indicated Detection status in Lane Keeping Assist. The name also states what the framework asks of the display: whether it is candid about what the system perceives. The report under evaluation is the lane-detection status shown to the driver by the production system. CANDID-LKA relates this status to two signals recorded from different sources during the same driving: the lane perception confidence of an independently implemented system observing the same road, and the driver's control response from the vehicle's Controller Area Network (CAN) bus. CANDID-LKA asks three questions of the displayed status. First, does it respond to the conditions that make lane markings harder to see? The \emph{internal test} answers this question by examining whether the status changes coherently with those conditions. Second, does it agree with other signals recorded at the same moment about the same road? The \emph{external test} answers this question by evaluating the status against the lane perception confidence and the driver's control response, treating degraded marking perceivability as an unobserved state that all three signals indicate, and estimating how completely the status reports it. Third, when the status is silent about degraded perception, what does it cost the driver? The same test answers this question by examining the outcomes of these silent seconds, quantified by the standard deviation of lateral position (SDLP) and the lateral excursion that follow. Because CANDID-LKA analyzes what a system reports about its own perception, and not lane detection itself, the same design applies to other driver-facing indicators, including availability reporting for adaptive cruise control and automatic emergency braking.

This study makes three contributions.

1. Applying CANDID-LKA to 39{,}800 seconds of naturalistic driving from 13 production vehicle models in the OpenLKA dataset \cite{wang2025openlka2}, we provide, to our knowledge, the first real-world evidence from production vehicles that the displayed lane-detection status is incomplete. Half of the sustained low-confidence episodes receive no \emph{not-detected} report. In silent seconds when the driver is not steering, the 3~s SDLP is about 60\% larger than after agreement seconds.

2. CANDID-LKA evaluates a proprietary driver-facing LKA report with a latent-class model that uses an independent vision system and the driver's steering signal, without access to the manufacturer's internal perception system or a ground-truth label. The framework could support consumer ratings, regulatory evaluation, and post-deployment audits, and extends to other driver-facing status indicators.

3. CANDID-LKA separates how often the status switches to \emph{not detected} from how long it stays there. This shows that the share of \emph{not-detected} time alone can hide long outages or frequent short ones, so evaluations should consider outage duration alongside how often outages start. Together with the large differences in how much each system reports, this suggests that the differences across manufacturers come partly from how each designs its status-reporting logic, and that more complete reporting is achievable.

The remainder of the paper reviews related work, formalizes the framework, describes the data and estimation, presents the two tests, and discusses their implications.

\section{Literature Review}\label{sec:related}

\subsection{Lane Keeping Performance}
Established standards define minimum functional and performance requirements for LKA, but no universally accepted procedure exists for comprehensive evaluation beyond regulatory compliance \cite{wei2024assessment}. Controlled testing therefore remains a primary approach to evaluating LKA performance. Track and instrumented-road studies show that lane detection is affected by pavement-marking retroreflectivity, contrast, and rainfall \cite{re2021testing}. Field programs extend such evaluations to adverse weather by staging scenarios involving worn markings and disappearing shoulders under rain and snow \cite{neumeister2019adverse,boyapati2023adverse}, and marking condition has likewise been used to assess infrastructure readiness for automated vehicles \cite{pike2024assessing}.

Simulation offers controlled access to rare and extreme conditions that are difficult to stage on the road. It has been used to model imperfect pavement-marking perception and the resulting lateral deviation \cite{yang2025simlkas}, to couple a real camera and a lane-segmentation algorithm to a high-fidelity vehicle model under rain and fog \cite{silva2026camera}, and to assess how lateral guidance systems behave with the driver in the loop \cite{wei2025simulator}.

Large-scale naturalistic datasets recently paired production vehicle signals with synchronized video and an independent onboard vision reference, enabling analysis of real-world lane-keeping performance at scale \cite{wang2025openlka2}, including a descriptive evaluation of perception performance across production systems \cite{alhuraish2025empirical}. This body of work establishes when and why lane detection degrades, on the road, in simulation, and at scale. It measures the sensing outcome, and does not address what the vehicle reports to the driver about that outcome.

\subsection{Perception Monitoring and Driver Trust}
Appropriate reliance on lane keeping assistance requires the system to signal when it can no longer see the lane, and the driver to act on that signal. These two requirements are complementary, but they have been studied separately, one on the system side and the other on the driver side.

On the system side, a body of work asks whether an automated driving system can recognize its own perception limits, a requirement identified as central to the safe deployment of learned perception \cite{muhammad2021deep}. Runtime monitors that predict misbehavior from a model's own confidence \cite{stocco2020misbehaviour}, out-of-distribution detectors \cite{henriksson2023ood}, and uncertainty estimators \cite{feng2022probabilistic} aim to flag when perception is unreliable so that the system can fall back safely. These methods assess perception from inside the software stack or against curated benchmarks, on research and prototype systems, and are distinct from the coarse detection status that production vehicles display.

On the driver side, human-factors research examines how drivers calibrate reliance on partial automation from the feedback the vehicle provides. Drivers form mental models of what the system is doing and seeing, and mis-calibrated trust and mode confusion are documented for Level-2 systems \cite{wilson2020trust,wischnewski2023trust}, with naturalistic evidence that regular users build reliance habits around system behavior and its cues \cite{mueller2024habits}. The displayed detection status is a primary cue for this reliance, yet whether it accurately reflects the system's perception, and what follows for the driver when it does not, has not been established.

\subsection{Multi-Source Reliability Estimation}
When no reference is available, the accuracy of a signal can be estimated by treating several imperfect signals as indicators of a common latent state and inferring, from their pattern of agreement, how often each is correct. This latent-class formulation originates in diagnostic testing, where the sensitivity and specificity of multiple imperfect tests are estimated jointly without a gold standard \cite{hui1980estimating}; the conditions under which the parameters are identifiable, and extensions that relax the assumption of independent errors, are well established \cite{jones2010identifiability,vacek1985dependence,dendukuri2001bayesian}, with reporting standards codified \cite{kostoulas2017stard}. In transportation safety, where unobserved heterogeneity across records is a central methodological concern \cite{mannering2016unobserved}, the same family of models is widely used, most often as latent class clustering that segments heterogeneous crash records into homogeneous accident types before relationships are estimated within each type \cite{depaire2008segmentation,deona2013latent}. Reference-free evaluation has also begun to appear in automated-driving perception, where an object detector's accuracy is estimated on unlabeled data from the consistency of its own predictions \cite{yoo2025automated}, though from a single model instead of the agreement of several independent signals.

This paper applies the diagnostic formulation to a driver-assistance system's own status reporting. Unlike crash-segmentation studies, where latent classes capture heterogeneity across records, the latent state here represents per-second lane-marking perceivability, and the inferential target is the accuracy of the observed signals. To our knowledge, reference-free latent-class estimation has not previously been applied to the self-reported output of a production driving-automation system.

\section{Analytical Framework}\label{sec:framework}

This section defines CANDID-LKA by describing the data it uses and the analyses its two tests conduct. For each driving second, the data come from two sources: the production vehicle, which records its displayed status and the driver's steering torque, and a third-party vision system, which observes the same road and reports its own perception confidence for the lane markings, its estimate of the vehicle's position in the lane, and forward video. The internal test relates the displayed status to driving conditions. The external test relates the displayed status to this lane perception confidence and to the driver's steering, estimating its trustworthiness with a latent-class model and three model-free measures. It then examines the silent seconds, in which the status reads \emph{detected} while the lane perception confidence is low: whether the driver is in control during them, and how the vehicle's lateral position varies afterwards when the driver is not.

\subsection{Variables}
Table~\ref{tab:variables} lists the variables CANDID-LKA uses for each driving second $t$. The displayed status $O_t$ is the object under evaluation. It is the detection state the system broadcasts to the driver-facing lane display, so it is a report about perception; what the system actually perceives stays internal to the system.

\begin{table}[!htbp]
\caption{Variables used by CANDID-LKA for each driving second $t$}\label{tab:variables}
\centering
\small
\begin{tabular}{p{0.07\textwidth}p{0.18\textwidth}p{0.44\textwidth}p{0.17\textwidth}}
\toprule
Symbol & Variable & Definition & Source \\
\midrule
$O_t$ & Displayed status & $1$ when the vehicle reports the lane as \emph{not detected}, $0$ when it reports \emph{detected} & Vehicle \\
$C_t$ & Lane perception confidence & $1$ when the third-party vision system's perception confidence for the lane line is low (below 0.5), $0$ otherwise & Third-party vision system \\
$I_t$ & Driver control response & $1$ when the driver's steering torque exceeds the vehicle's 95th percentile within the following 3~s & Vehicle \\
$H_t$ & Driver in control & $1$ when the driver's mean steering torque over the following 3~s exceeds the vehicle's median per-second torque & Vehicle \\
$Y_t$ & Lateral position & Position of the vehicle in the lane estimated by the vision system at 10~Hz; the 3~s SDLP and lateral excursion are computed from it & Third-party vision system \\
$X_t$ & Driving conditions & Marking condition (clear, low-contrast, faded, or occluded), lighting, surface, road segment, traffic, lead-vehicle presence, speed, and road curvature & Scene labels from the video; vehicle; vision system \\
$F_t$ & Silent second & $1$ when $O_t = 0$ and $C_t = 1$ (status \emph{detected} while the perception confidence is low), $0$ otherwise & Derived from $O_t$ and $C_t$ \\
$D_t$ & Ground-truth perceivability & $1$ when lane-marking perceivability is degraded, $0$ otherwise & Not observable; estimated by the latent-class model \\
\bottomrule
\end{tabular}
\end{table}

\subsection{Internal Test}
The internal test evaluates whether the displayed status responds to driving conditions that reduce lane-marking perceivability. It works in two steps. The first step gives the overall picture: under each condition, how often does the status report \emph{not detected}? This answers the question directly, because a status that tracks marking difficulty should report \emph{not detected} more often as markings become harder to see. The second step looks at the dynamics behind that share: how often the status switches to \emph{not detected}, and how quickly it switches back. This step is needed because the share alone can hide how the status behaves, since the same share can come from frequent short losses or from rare long ones.

The first step uses the condition-response function
\begin{equation}\label{eq:response}
p(x) \;=\; \Pr\left(O_t = 1 \mid X_t = x\right),
\end{equation}
the share of seconds under condition $x$ in which the status reports \emph{not detected}. The effect of a condition is expressed as an odds ratio, the ratio of the odds of a \emph{not-detected} report under that condition to the odds under a reference condition $x_0$,
\begin{equation}\label{eq:or}
\mathrm{OR}(x) \;=\; \frac{p(x)\,/\,[1-p(x)]}{p(x_0)\,/\,[1-p(x_0)]}.
\end{equation}
Both are estimated with one logistic regression of the status on the driving conditions, fit to all systems together and then to each make separately to examine whether the makes respond differently to a condition. Model fit is reported using the area under the receiver operating characteristic curve (AUC) and McFadden's pseudo-$R^2$.

The second step describes how that share arises, through two rates: the loss rate $\lambda_{\text{loss}}(x)$, the probability that the status switches from \emph{detected} to \emph{not detected} in the next second, and the recovery rate $\lambda_{\text{rec}}(x)$, the probability that it switches back,
\begin{equation}\label{eq:rates}
\lambda_{\text{loss}}(x) = \Pr\!\left(O_t = 1 \mid O_{t-1} = 0,\, X_t = x\right), \qquad
\lambda_{\text{rec}}(x) = \Pr\!\left(O_t = 0 \mid O_{t-1} = 1,\, X_t = x\right).
\end{equation}
Multiplied by 3{,}600, each gives the number of switches per hour spent in the starting state. The effect of a condition on either rate is expressed as a rate ratio, $\lambda(x)/\lambda(x_0)$. A responsive status has $p(x)$ and $\lambda_{\text{loss}}(x)$ increasing as $x$ moves toward conditions that degrade markings.

The three quantities are linked. In steady state,
\begin{equation}\label{eq:steadystate}
p(x) \;=\; \frac{\lambda_{\text{loss}}(x)}{\lambda_{\text{loss}}(x) + \lambda_{\text{rec}}(x)} .
\end{equation}
The share $p(x)$ therefore combines two behaviors: how often the status switches to \emph{not detected}, and how long it stays there. A maximal run of consecutive \emph{not-detected} seconds is a reported \emph{outage}, with mean duration $1/\lambda_{\text{rec}}(x)$. Two systems with the same $p(x)$ can differ entirely in these two behaviors, and $p(x)$ can even fall as conditions worsen if losses become more frequent but shorter. The two rates are estimated from the switches by Poisson regression, for all systems together and for each make. Because switches are few, these models include only the marking condition.

\subsection{External Test}
The internal test uses only the displayed status and driving conditions. A status can respond to conditions on average yet still misreport at a given moment. The external test therefore evaluates the displayed status against two other signals recorded during the same driving: the perception confidence $C$ of an independent system observing the same road, and the driver control response $I$. Its central quantity is the trustworthiness of the status, estimated by a latent-class model that combines all three signals. Three model-free measures, the agreement between the status and the perception confidence, the episode miss rate, and the driver-response risk ratio, support that estimate; each is computed for the full sample and by make, with confidence intervals from a route-level bootstrap (2{,}000 resamples). The external test then takes the silent seconds, those in which the status reads \emph{detected} while the perception confidence is low, and asks where they occur and what follows them.

\textbf{Trustworthiness.} The displayed status, the perception confidence, and the driver control response are treated as three imperfect indicators of an unobserved binary state $D_t$, equal to $1$ when lane-marking perceivability is degraded. Each indicator $k \in \{O, C, I\}$ has a sensitivity $Se_k = \Pr(S_{kt}=1 \mid D_t=1)$, the probability that it equals 1 when perceivability is degraded, and a specificity $Sp_k = \Pr(S_{kt}=0 \mid D_t=0)$, the probability that it equals 0 when perceivability is good, where $S_{kt}$ is the value of indicator $k$ in second $t$. The prevalence $\pi = \Pr(D_t=1)$ is the share of seconds in which perceivability is degraded. The probability of an observed pattern $s = (s_O, s_C, s_I)$ is
\begin{equation}\label{eq:lcm}
\Pr(s) \;=\; \pi \prod_{k} Se_k^{\,s_k}(1 - Se_k)^{1-s_k} \;+\; (1-\pi)\prod_{k}(1 - Sp_k)^{s_k} Sp_k^{\,1-s_k} .
\end{equation}
The model is estimated from the observed pattern of agreement among the three signals \cite{hui1980estimating}, and yields the sensitivity and specificity of the displayed status directly: the share of degraded seconds in which it reports \emph{not detected}, and the share of other seconds in which it reports \emph{detected}. The model designates no signal as a reference; each of the three can be wrong, and their errors are assumed to be independent given the true state. It is fit by maximum likelihood with the expectation-maximization algorithm from 10 random starts, for all systems together and for each make.

\textbf{Agreement.} How closely the displayed status and the perception confidence agree about the same road, summarized by Cohen's kappa,
\begin{equation}\label{eq:kappa}
\kappa \;=\; \frac{\Pr(O_t = C_t) - P_e}{1 - P_e},
\end{equation}
where $P_e$ is the agreement two unrelated signals would reach given how often each reports loss. Thus $\kappa = 0$ when the status and the perception confidence agree no more than two unrelated signals would, and $\kappa = 1$ when they agree perfectly.

\textbf{Episode miss rate.} Whether the displayed status warns the driver during periods of low perception confidence. A driver relies on the status to signal when the lane becomes hard to see, so the status should report \emph{not detected} at some point during such a period. A low-confidence episode is a stretch of consecutive seconds with $C_t = 1$, from the second the perception confidence falls below 0.5 until it recovers. Only episodes lasting at least two seconds are counted. The episode miss rate is the share of these episodes in which the status never reports \emph{not detected}.

\textbf{Driver-response risk ratio.} Whether the displayed status corresponds to the driver's control response. If the status conveys real information, a \emph{not-detected} report should be followed more often by a corrective input than a \emph{detected} report. The driver-response risk ratio is the ratio of these probabilities,
\begin{equation}\label{eq:rr}
\mathrm{RR} \;=\; \frac{\Pr(I_t = 1 \mid O_t = 1)}{\Pr(I_t = 1 \mid O_t = 0)} ,
\end{equation}
which equals $1$ when the status carries no information about the driver's response. A high-torque input can have several causes, among them the steering a curve requires, a lane change, and a correction of the lane-keeping function. The risk ratio therefore does not by itself say why the driver steered; the silence outcome analysis examines where these inputs occur.

\textbf{Silence outcome analysis.} This analysis asks what happens to the vehicle's position in the lane after a silent second. It uses the silent seconds of the external-test sample, in which the status reads \emph{detected} while the perception confidence is low ($O_t = 0$, $C_t = 1$), and compares them with agreement seconds, in which the status reads \emph{detected} and the perception confidence is not low ($O_t = 0$, $C_t = 0$). Both are split by whether the driver is in control ($H_t$). The driver is defined as in control when the mean steering torque over the following 3~s is above the vehicle's median per-second torque. $H_t$ is different from $I_t$: $I_t$ marks a rare, hard input and is used in the latent-class model, while $H_t$ says who is steering.

In each group, two metrics are computed over the following 3~s: the 3~s SDLP, which is the standard deviation of $Y_t$, and the lateral excursion, which is the largest minus the smallest $Y_t$. Each metric is compared between silent and agreement seconds with a linear regression that adjusts for speed, planned lateral acceleration (log and squared terms), and vehicle model, with standard errors clustered by route. The planned lateral acceleration is the curvature of the path planned by the vision system times the speed squared, and it accounts for the steering the road requires. Because the not-in-control seconds are the moments with the most potential risk, the comparison for this group is further broken down by marking condition, lighting, and road type, and repeated on straight road only.

\section{Data Description}\label{sec:data}
The data come from OpenLKA-Failure, the subset of the OpenLKA naturalistic dataset \cite{wang2025openlka2} that collects driving in which lane keeping is difficult. This subset comprises 4{,}947 one-minute clips (about 82 hours) recorded with a windshield-mounted comma~3X device. Each clip carries three synchronized channels: the vehicle's CAN bus, decoded through comma.ai's reverse-engineered opendbc message definitions; the openpilot vision model's per-lane-line confidence, lane geometry, and lane-change state \cite{commaai2024openpilot}; and scene labels, extracted by a vision-language model (VLM) from simultaneously recorded forward video. The recordings combine a research campaign in rental vehicles around Tampa, Florida in mid-2024 with comma.ai community-contributed drives across the United States.

Of the 41 vehicle models in the dataset, 16 models from three manufacturers broadcast the displayed status, the lane-detection status shown to the driver, in a form that can be decoded. The screening described below removes three of them (the Toyota Alphard, the Honda Civic, and the Toyota Corolla), and the remaining 13 models form the analysis sample. Toyota and Lexus report a discrete display state, Hyundai an ordinal confidence level, and Honda a continuous probability. The displayed status is encoded as follows:
\begin{itemize}[nosep]
\item Toyota and Lexus: the display state takes four values; states 1 and 3 mark the lane as \emph{detected} and state 2 as \emph{not detected}, while state 0 appears only when lane keeping is inactive and is excluded from analysis.
\item Hyundai: the lane-line confidence ranges from 0 to 3, and a value of 0 marks \emph{not detected}.
\item Honda: the lane-line probability is continuous, and a value below 0.5 marks \emph{not detected}.
\end{itemize}
The analysis keeps the seconds in which the displayed status comes from the vehicle's own system (not from openpilot), the status is active, the vehicle moves faster than 3~m/s, and no lane change is under way at that second. The data are further screened as follows:
\begin{itemize}[nosep]
\item duplicate clips are counted once;
\item the Toyota Alphard is removed because its status never changes;
\item the Honda Civic is removed because its left-line status cannot be decoded;
\item only files logged at 100~Hz are kept, because the status in 10~Hz files is interpolated; this removes the Toyota Corolla;
\item clips whose factory and vision data are not time-aligned are removed; and
\item seconds with signs that the status was written by openpilot are removed.
\end{itemize}
After screening, the sample comprises 39{,}800 seconds across 853 clips and 89 routes: 385 Toyota and Lexus clips from 8 models (16{,}605~s), 388 Hyundai clips from 4 models (18{,}744~s), and 80 Honda clips from 1 model (4{,}451~s).

\section{Results and Discussion}\label{sec:results}

\subsection{Response of the Displayed Status to Driving Conditions}
Table~\ref{tab:cond} reports the odds ratios for all systems together and for each make. Across all systems, occlusion has the largest odds ratio, 1.94: with the other conditions held the same, the status reports \emph{not detected} in about 30\% of seconds on occluded markings, compared with 20\% on clear markings. However, this difference is not statistically significant, because its 95\% confidence interval (0.79 to 4.76) includes 1. The other marking conditions, low-contrast and faded, show little association with the status. Among the other conditions, the presence of a lead vehicle and higher speed are significantly associated with fewer \emph{not-detected} reports. With a lead vehicle present, the status reports \emph{not detected} in about 17\% of seconds, compared with 24\% without one (odds ratio 0.56). When speed is one standard deviation (about 30~km/h) higher than observed, the share falls from 20\% to 13\% (odds ratio 0.52). Night driving shows a similar drop, from 21\% of seconds in daylight to 12\% at night (odds ratio 0.42), but this difference is also not statistically significant. Road curvature enters the model as two terms, the logarithm of curvature and its square. Across all systems, neither is significant (odds ratio 1.02, 95\% CI 0.94 to 1.11, and odds ratio 1.00, 95\% CI 0.98 to 1.02), so sharper curves do not change how often the status reports \emph{not detected}. Overall, \emph{not-detected} reports do not consistently become more likely under conditions that make lane markings harder to see.

To examine why, we fit a model that uses only the vehicle model as the variable, and its AUC is 0.73, while including all the driving conditions raises the AUC only to 0.80. This shows that most of the variation in the displayed status comes from which vehicle it is, not from the driving conditions. The next analysis therefore examines whether the effects of marking conditions differ across makes.

\begin{table}[!htbp]
\caption{Odds ratios for a \emph{not-detected} report under different driving conditions}\label{tab:cond}
\centering
\footnotesize
\setlength{\tabcolsep}{2.5pt}
\begin{tabular}{lccccc}
\toprule
 & \multicolumn{4}{c}{Displayed status (\emph{not detected})} & Perception \\
\cmidrule(lr){2-5}
Condition & All systems & Toyota & Hyundai & Honda & confidence (low) \\
\midrule
\multicolumn{6}{l}{\emph{Marking condition (vs.\ clear)}} \\
\quad Low-contrast & 0.79 (0.48--1.29) & 0.81 (0.42--1.58) & 0.67 (0.29--1.56) & 1.92 (1.16--3.16)$^{*}$ & 1.31 (0.98--1.75) \\
\quad Faded & 0.99 (0.53--1.87) & 0.55 (0.30--1.00) & 2.11 (1.07--4.16)$^{*}$ & 2.60 (1.17--5.78)$^{*}$ & 1.31 (0.79--2.19) \\
\quad Occluded & 1.94 (0.79--4.76) & 3.54 (1.09--11.42)$^{*}$ & 1.54 (0.34--6.92) & 0.38 (0.23--0.62)$^{*}$ & 0.82 (0.34--2.00) \\
\addlinespace
\multicolumn{6}{l}{\emph{Other conditions}} \\
\quad Night & 0.42 (0.16--1.16) & 0.54 (0.18--1.59) & 0.16 (0.07--0.36)$^{*}$ & 0.02 (0.01--0.03)$^{*\dagger}$ & 0.77 (0.45--1.30) \\
\quad Wet surface & 2.63 (0.67--10.34) & 1.10 (0.20--6.00) & 8.96 (0.77--103.67) & -- & 1.18 (0.72--1.95) \\
\quad Intersection & 1.22 (0.81--1.85) & 1.23 (0.67--2.25) & 1.06 (0.47--2.38) & 1.51 (0.64--3.59) & 1.26 (0.88--1.80) \\
\quad Free flow & 1.47 (0.75--2.89) & 1.76 (0.66--4.75) & 1.07 (0.44--2.60) & 1.83 (1.02--3.28)$^{*}$ & 1.62 (1.09--2.41)$^{*}$ \\
\quad Lead vehicle & 0.56 (0.33--0.96)$^{*}$ & 0.55 (0.26--1.14) & 0.66 (0.27--1.60) & 0.35 (0.14--0.83)$^{*}$ & 0.40 (0.24--0.65)$^{*}$ \\
\quad Speed (+1 SD) & 0.52 (0.39--0.70)$^{*}$ & 0.59 (0.42--0.84)$^{*}$ & 0.48 (0.28--0.83)$^{*}$ & 0.35 (0.24--0.51)$^{*}$ & 0.15 (0.12--0.18)$^{*}$ \\
\quad Log curvature & 1.02 (0.94--1.11) & 1.02 (0.91--1.15) & 1.03 (0.91--1.16) & 1.76 (1.25--2.48)$^{*}$ & 1.49 (1.39--1.60)$^{*}$ \\
\quad Log curvature$^2$ & 1.00 (0.98--1.02) & 1.00 (0.98--1.02) & 1.00 (0.97--1.03) & 1.09 (1.02--1.17)$^{*}$ & 1.13 (1.11--1.15)$^{*}$ \\
\bottomrule
\multicolumn{6}{p{0.97\textwidth}}{\footnotesize Route-clustered 95\% CI; $^{*}$ $p<0.05$. $^{\dagger}$Honda has 1 \emph{not-detected} second out of 1{,}475 night seconds. --: Honda has no wet seconds. All-systems model: AUC 0.80 (0.73 with vehicle-model fixed effects alone); McFadden pseudo-$R^2$ 0.20.}
\end{tabular}
\end{table}

Fitting each make separately (Table~\ref{tab:cond}) shows that the makes respond differently. On faded markings, Hyundai and Honda report \emph{not detected} more often than on clear markings: about 23\% of seconds against 14\% for Hyundai (odds ratio 2.11, 95\% CI 1.07 to 4.16), and 9\% against 5\% for Honda (2.60, 1.17 to 5.78). Toyota reports it less often, 21\% against 31\% (0.55, 0.30 to 1.00). Honda also reports \emph{not detected} more often on low-contrast markings (1.92, 1.16 to 3.16). On occluded markings, Toyota reports \emph{not detected} more often, in 55\% of seconds against 31\% (3.54, 1.09 to 11.42). Hyundai's odds ratio there is also above 1 but not statistically significant, and Honda reports it less often (0.38, 0.23 to 0.62). The other conditions act the same way in every make: higher speed, night, and a lead vehicle each come with fewer \emph{not-detected} reports, although not every estimate is significant. Road curvature is associated with the status only for Honda, where both curvature terms are significant. The perception confidence, by contrast, is low more often on every degraded marking than on clear markings: in 18\% of seconds on low-contrast markings, 13\% on faded, and 24\% on occluded, against 8\% on clear. After adjusting for the other conditions, its odds ratios on low-contrast and faded markings are above 1 (1.31 for both), but not statistically significant. Taken together, the results show that the displayed status does not respond to marking degradation in the same way across makes. The next subsection examines whether this inconsistency arises from differences in how the status switches to \emph{not detected} and back.

\subsection{Switching Dynamics of the Displayed Status}
Table~\ref{tab:dynamics} reports the loss and recovery rate ratios by marking condition, for all systems together and for each make. In what follows, an outage is a period in which the status stays \emph{not detected}.

Across all systems, occlusion is the only condition with a clear effect. On occluded markings, the status switches to \emph{not detected} about four times as often as on clear markings (loss rate ratio 4.10, 95\% CI 1.95 to 8.60). Once it reads \emph{not detected}, it switches back about 1.8 times as fast (recovery rate ratio 1.84, 0.78 to 4.33), although this second difference is not statistically significant. Occluded markings therefore produce outages that are frequent and comparatively short. Across all systems, low-contrast and faded markings show little association with either rate, and on faded markings the recovery rate moves in opposite directions for different makes.

\begin{table}[!htbp]
\caption{Loss and recovery rate ratios of the displayed status by marking condition}\label{tab:dynamics}
\centering
\small
\setlength{\tabcolsep}{5pt}
\begin{tabular}{llccc}
\toprule
 & Transition & Low-contrast & Faded & Occluded \\
\midrule
All systems & Loss & 1.44 (0.94--2.21) & 1.47 (0.85--2.54) & 4.10 (1.95--8.60)$^{*}$ \\
 & Recovery & 1.12 (0.55--2.27) & 1.18 (0.59--2.39) & 1.84 (0.78--4.33) \\
\addlinespace
Toyota & Loss & 1.01 (0.54--1.87) & 0.95 (0.43--2.06) & 8.10 (3.73--17.60)$^{*}$ \\
 & Recovery & 1.04 (0.29--3.69) & 1.57 (0.89--2.78) & 2.77 (0.92--8.31) \\
\addlinespace
Hyundai & Loss & 1.51 (0.72--3.15) & 1.50 (0.54--4.20) & 3.61 (1.75--7.44)$^{*}$ \\
 & Recovery & 0.87 (0.40--1.90) & 0.46 (0.13--1.70) & 1.44 (0.35--5.84) \\
\addlinespace
Honda & Loss & 2.34 (1.24--4.43)$^{*}$ & 2.33 (0.73--7.44) & --- \\
 & Recovery & 1.06 (0.53--2.14) & 1.67 (0.97--2.89) & 0.64 (0.44--0.93)$^{*}$ \\
\bottomrule
\multicolumn{5}{p{0.92\textwidth}}{\footnotesize Rate ratios vs.\ clear markings from Poisson regressions on the marking condition; route-clustered 95\% CI; $^{*}$ $p<0.05$. Switches are few, so the models include only the marking condition. ---: no loss began under occlusion for Honda.}
\end{tabular}
\end{table}

The three makes behave differently. For Toyota, occlusion makes the status switch to \emph{not detected} about eight times as often as on clear markings (loss rate ratio 8.10, 95\% CI 3.73 to 17.60), and the mean observed outage shortens by half, to about 12~s. For Hyundai, outages on faded markings last longer: the median observed outage rises from 5~s on clear markings to 14~s on faded ones, and 63\% of its faded-marking outages are still ongoing when the clip ends, against 49\% on clear markings. On faded markings, Hyundai's status switches to \emph{not detected} 1.5 times as often (loss rate ratio 1.50, 0.54 to 4.20) and switches back about half as fast (recovery rate ratio 0.46, 0.13 to 1.70), but neither difference is statistically significant, so this pattern rests mainly on the outage durations. For Honda, outages are short under every marking condition (median 3 to 4~s), but they start more often on low-contrast markings (2.34, 1.24 to 4.43) and on faded markings (2.33, 0.73 to 7.44, not statistically significant).

These dynamics explain why the share of time in the \emph{not-detected} state can move against marking difficulty. The share depends on both rates (Equation~\ref{eq:steadystate}), so a system that switches back faster spends less time reporting \emph{not detected}, even if it switches to \emph{not detected} just as often. Toyota on faded markings is this case: it switches to \emph{not detected} about as often as on clear markings (loss rate ratio 0.95, 0.43 to 2.06), but switches back faster (recovery rate ratio 1.57, 0.89 to 2.78, not statistically significant), and its share of \emph{not-detected} seconds falls from 31\% to 21\%. The share of \emph{not-detected} time therefore cannot be read as a measure of marking difficulty, and whether the displayed status can be trusted has to be judged against evidence from outside the system.

\subsection{Trustworthiness of the Displayed Status}
\begin{figure}[!htbp]
\centering
\includegraphics[width=0.98\textwidth]{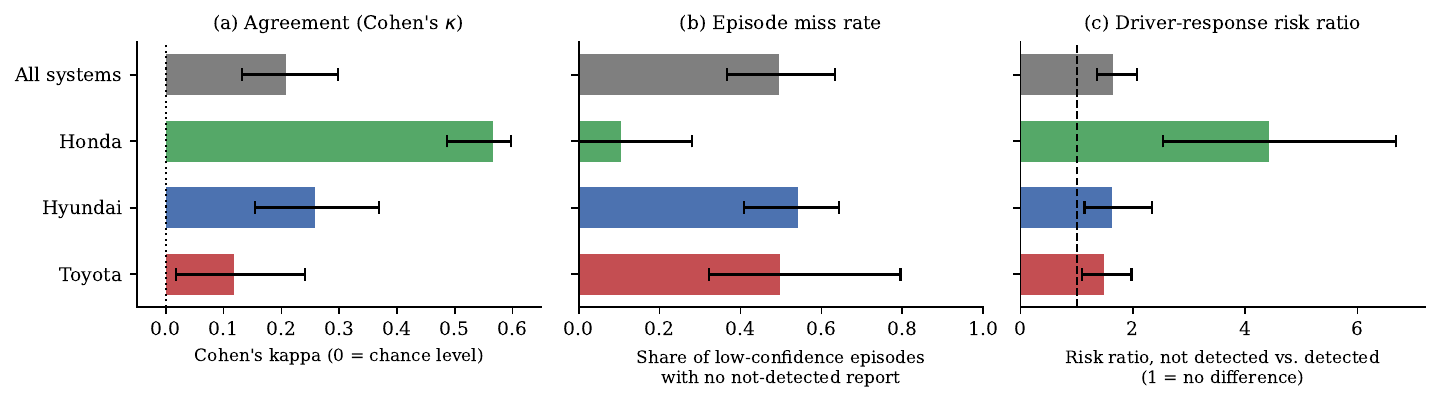}
\caption{Agreement, episode miss rate, and driver-response risk ratio, by make (route-bootstrap 95\% CI)}\label{fig:trust}
\end{figure}

The external test compares the displayed status with two other signals, the perception confidence and the driver's control response. Because the driver control response looks 3~s ahead, the last 3~s of each clip are dropped, so the external test uses 37{,}353 seconds across 825 clips. From the observed combinations of the three signals, the latent-class model estimates the share of seconds with degraded perceivability and the sensitivity and specificity of each signal (Table~\ref{tab:trust}). Three measures computed directly from the signals support these estimates (Figure~\ref{fig:trust}).

\begin{table}[!htbp]
\caption{Latent-class estimates of prevalence, sensitivity, and specificity, by make}\label{tab:trust}
\centering
\footnotesize
\setlength{\tabcolsep}{4pt}
\begin{tabular}{lcccc}
\toprule
 & All systems & Toyota & Hyundai & Honda \\
\midrule
Prevalence of degraded perceivability & 0.13 (0.08--0.19) & 0.14 (0.09--0.22) & 0.14 (0.07--0.29) & 0.05 (0.01--0.13) \\
\addlinespace
\multicolumn{5}{l}{\emph{Displayed status}} \\
\quad Sensitivity & 0.47 (0.34--0.62) & 0.46 (0.20--0.67) & 0.44 (0.33--0.63) & 0.79 (0.75--0.89) \\
\quad Specificity & 0.84 (0.74--0.92) & 0.75 (0.56--0.92) & 0.89 (0.83--0.94) & 0.98 (0.94--0.99) \\
\quad Positive predictive value & 0.30 (0.22--0.44) & 0.24 (0.15--0.33) & 0.40 (0.23--0.67) & 0.67 (0.50--0.84) \\
\quad Negative predictive value & 0.92 (0.85--0.95) & 0.89 (0.80--0.94) & 0.91 (0.79--0.97) & 0.99 (0.97--1.00) \\
\addlinespace
\multicolumn{5}{l}{\emph{Perception confidence}} \\
\quad Sensitivity & 0.80 (0.59--1.00) & 0.79 (0.50--1.00) & 0.73 (0.42--1.00) & 0.72 (0.58--0.94) \\
\quad Specificity & 0.99 (0.97--1.00) & 0.98 (0.91--1.00) & 1.00 (0.96--1.00) & 1.00 (0.99--1.00) \\
\addlinespace
\multicolumn{5}{l}{\emph{Driver control response}} \\
\quad Sensitivity & 0.29 (0.25--0.36) & 0.35 (0.32--0.52) & 0.24 (0.19--0.35) & 0.43 (0.19--0.65) \\
\quad Specificity & 0.94 (0.93--0.95) & 0.95 (0.94--0.99) & 0.94 (0.91--0.96) & 0.93 (0.90--0.95) \\
\bottomrule
\multicolumn{5}{p{0.93\textwidth}}{\footnotesize Route-bootstrap 95\% CIs (2{,}000 resamples).}
\end{tabular}
\end{table}

\textbf{Trustworthiness.} The latent-class model estimates that 13\% of seconds have degraded perceivability (the prevalence in Table~\ref{tab:trust}). This latent state closely follows the perception confidence. The perception confidence is low in 80\% of the degraded seconds (its sensitivity, 0.80) and almost never low in the other seconds (its specificity, 0.99). This supports reading the latent state as degraded perceivability, and using the perception confidence as the reference for the analyses that follow.

Against this state, the displayed status reports only a minority. Its sensitivity, the share of degraded seconds in which it reports \emph{not detected}, is 0.47 (95\% CI 0.34 to 0.62) across all systems. Its specificity, the share of other seconds in which it reports \emph{detected}, is 0.84. By make, the sensitivity is 0.46 for Toyota, 0.44 for Hyundai, and 0.79 for Honda. So Toyota's and Hyundai's status report fewer than half of the degraded seconds, while Honda's reports about four in five. A \emph{not-detected} report is also a weak signal on its own: only 30\% of \emph{not-detected} reports fall in degraded seconds (the positive predictive value). A \emph{detected} report is more reliable: 92\% of \emph{detected} reports fall in seconds that are not degraded (the negative predictive value). Because the data come from OpenLKA-Failure, the subset of OpenLKA that collects difficult driving, these predictive values describe difficult driving; in routine driving, where degraded seconds are rarer, the share of \emph{not-detected} reports that fall in degraded seconds would be lower.

The driver's control response catches few degraded seconds (sensitivity 0.24 to 0.43), but it rarely occurs in seconds that are not degraded (specificity about 0.94). Much of it is steering the road requires, as the next subsection shows. The status sensitivity from the model (0.47) is close to the share of low-confidence seconds in which the status reports \emph{not detected}, 0.45 (1{,}818 of 4{,}054 low-confidence seconds), which is computed directly from the signals. So the estimate does not depend on the driver signal.

\textbf{Agreement.} Agreement between the status and the perception confidence (Cohen's kappa, where 0 is chance level and 1 is perfect agreement) is 0.21 (95\% CI 0.13 to 0.30) across all systems, which means that, on average, the two signals agree only a little more often than chance would produce. The agreement differs widely by make: it is 0.12 (0.02 to 0.24) for Toyota, 0.26 (0.15 to 0.37) for Hyundai, and 0.57 (0.49 to 0.60) for Honda. In other words, Honda's status agrees with the perception confidence far more than Toyota's or Hyundai's does.

\textbf{Episode miss rate.} In half of the low-confidence episodes, which are periods of at least 2~s in which the perception confidence stays low, the status never reports \emph{not detected} at any point. This share is 50\% for Toyota and 54\% for Hyundai, but only 10\% for Honda. When the status does report \emph{not detected} during an episode, it does so right away: the median time from the start of the episode to the first \emph{not-detected} report is 0~s (1~s for Honda). This shows that the status does not tend to report \emph{not detected} late; when it fails, it never reports \emph{not detected} at all. Taken over all driving time, the status reads \emph{detected} while the perception confidence is low for 216 seconds of every driving hour across all systems, with 271 for Toyota, 209 for Hyundai, and 37 for Honda.

\textbf{Driver-response risk ratio.} After a \emph{not-detected} report, a high-torque steering input follows within 3~s more often than after a \emph{detected} report. Across all systems it follows 1.65 times as often (95\% CI 1.37 to 2.07), and by make the ratio is 1.49 for Toyota (1.09 to 1.98), 1.63 for Hyundai (1.14 to 2.34), and 4.43 for Honda (2.53 to 6.68). The difference is even larger after silent seconds, in which the status reads \emph{detected} while the perception confidence is low. On every make, a high-torque input follows a silent second 3.6 to 5.3 times as often as it follows a second in which both signals say the lane is seen (27\% against 6\% across all systems). This gap remains after adjusting for marking condition, lighting, surface, lead-vehicle presence, speed, and vehicle model, with the odds still 2.8 to 6.0 times higher. However, once the lateral acceleration that the planned path requires over the same 3~s is also taken into account, the ratio falls to 0.89 (0.62 to 1.27) across all systems, and the ratio after a \emph{not-detected} report falls to 0.83 (0.67 to 1.02); neither is statistically significant. This shows that the driver's steering follows what the road requires, not what the status reports.

The four measures rest on different foundations. Agreement, the episode miss rate, and the risk ratios are computed directly from the signals, while the sensitivity, specificity, and predictive values come from the latent-class model. Even so, they lead to the same conclusion: in about half of the periods when the perception confidence is low, the displayed status still reads \emph{detected}. The next subsection examines where these silent seconds occur and what follows them.

\subsection{Outcomes of the Silent Seconds}\label{sec:silence}
The silent seconds, in which the status reads \emph{detected} while the perception confidence is low, are moments of potential risk, and they occur 216 times per driving hour across all systems in the dataset. To examine what these silent seconds cost, we analyze them further. The silent seconds are first split into two groups: those in which the driver is in control, and those in which the driver is not. The latter are generally the riskier ones, because the lane keeping is then left to the system while the status gives no warning. To set the threshold between the two groups, we use Toyota's log, which records whether the lane keeping function is steering the vehicle. While the lane keeping function is steering, the driver applies little torque (quartiles of 17, 30, and 79, in raw units); while the lane keeping function is not steering, the driver does the steering and applies much more (85, 135, and 189). The vehicle's median per-second torque falls between these two ranges, so it separates the two situations well. We therefore use it as the threshold for grouping: a silent second is grouped as in control when the driver's mean torque over the following 3~s is above the vehicle's median, and as not in control otherwise.

Table~\ref{tab:control} shows the result of the split. The driver is in control in 78\% of silent seconds, compared with 48\% of agreement seconds. To see what the vehicle is doing in these seconds, each 3~s window is classed as a lane change (the lane-center reading jumps by more than 1~m), a sharp turn (planned lateral acceleration above 1~m/s$^2$), a gentle curve (0.5 to 1~m/s$^2$), or straight road (below 0.5~m/s$^2$). The in-control silent seconds are mostly sharp turns (65\%), at a median speed of 29~km/h, with straight road (18\%) and gentle curves (16\%) making up the rest. In contrast, the in-control agreement seconds are mostly on straight road (56\%) at 88~km/h. This accounts for most of the reported four to five times higher driver response after silent seconds: the high-torque inputs counted by $I_t$ are mostly the driver steering through a turn. The lateral position also varies more after silent seconds in this group: the adjusted 3~s SDLP is 5.2~cm higher (95\% CI 3.3 to 7.1) than after in-control agreement seconds, and the lateral excursion exceeds 50~cm in 75\% of windows against 22\% (Table~\ref{tab:control}). This is reasonable, because most of these silent seconds are sharp turns, where the vehicle moves more within the lane than on the straight roads that make up most of the in-control agreement seconds.

\begin{table}[!htbp]
\caption{Silent and agreement seconds by whether the driver is in control over the following 3~s, all systems}\label{tab:control}
\centering
\small
\setlength{\tabcolsep}{4pt}
\begin{tabular}{lcccc}
\toprule
 & \multicolumn{2}{c}{Silent seconds} & \multicolumn{2}{c}{Agreement seconds} \\
\cmidrule(lr){2-3}\cmidrule(lr){4-5}
 & In control & Not in control & In control & Not in control \\
\midrule
Share of seconds & 78\% & 22\% & 48\% & 52\% \\
Seconds per driving hour & 166 & 48 & & \\
Median speed (km/h) & 29 & 36 & 88 & 92 \\
Maneuver: sharp turn & 65\% & 19\% & 19\% & 4\% \\
\quad gentle curve & 16\% & 16\% & 22\% & 13\% \\
\quad straight road & 18\% & 64\% & 56\% & 83\% \\
\quad lane change & 2\% & 0\% & 3\% & 1\% \\
High-torque input within 3~s ($I_t$) & 35\% & 0\% & 13\% & 0\% \\
\addlinespace
\multicolumn{5}{l}{\emph{Lateral position over the following 3~s}} \\
\quad 3~s SDLP, median (cm) & 22.3 & 12.2 & 7.6 & 4.2 \\
\quad Excursion above 50~cm & 75\% & 46\% & 22\% & 6\% \\
\quad Adjusted SDLP difference (cm)$^{a}$ & 5.2 (3.3--7.1) & 5.4 (3.8--7.0) & & \\
\bottomrule
\multicolumn{5}{p{0.9\textwidth}}{\footnotesize $^{a}$Silent minus agreement seconds in the same group, adjusted for speed, planned lateral acceleration, and vehicle model; route-clustered 95\% CI.}
\end{tabular}
\end{table}

The remaining 22\% of silent seconds, 48 per driving hour, are the cases in which the driver is not in control, so the lane keeping is left to the system while the status gives no warning. Two thirds of them are on straight road, at a median speed of 36~km/h. For these seconds, the comparison is with agreement seconds in which the driver is also not in control, and the lateral position varies much more after a silent second. The median 3~s SDLP is 12.2~cm after a silent second, compared with 4.2~cm after an agreement second, and the adjusted difference is $+5.4$~cm (95\% CI 3.8 to 7.0), about 60\% more. The lateral excursion within 3~s exceeds 50~cm in 46\% of the windows after a silent second, compared with 6\% after an agreement second. Table~\ref{tab:openroad} breaks this result down by driving condition. On straight road alone, the difference is even larger: the SDLP is 6.4~cm higher (5.0 to 7.8), and the excursion exceeds 50~cm in 36\% of windows against 3\%.

\begin{table}[!htbp]
\caption{Silence with the driver not in control: the 3~s SDLP and lateral excursion that follow, by driving condition, all systems}\label{tab:openroad}
\centering
\footnotesize
\setlength{\tabcolsep}{3pt}
\begin{tabular}{lcccc}
\toprule
 & Silent & 3~s SDLP, median (cm) & SDLP adjusted & Excursion above 50~cm \\
Condition & seconds & agreement / silent & difference (cm) & agreement / silent \\
\midrule
All not-in-control seconds & 500 & 4.2 / 12.2 & 5.4 (3.8--7.0) & 6\% / 46\% \\
Straight road only & 322 & 3.8 / 10.7 & 6.4 (5.0--7.8) & 3\% / 36\% \\
\addlinespace
Clear markings & 187 & 4.1 / 11.8 & 5.5 (3.6--7.3) & 6\% / 44\% \\
Low-contrast markings & 154 & 4.6 / 11.9 & 6.3 (3.5--9.0) & 6\% / 44\% \\
Faded markings & 101 & 4.1 / 13.9 & 4.1 (0.8--7.4) & 8\% / 47\% \\
\addlinespace
Day & 410 & 4.2 / 12.5 & 5.5 (3.9--7.2) & 7\% / 46\% \\
Night & 61 & 4.2 / 10.1 & 4.9 (1.9--7.9) & 4\% / 31\% \\
\addlinespace
Highway & 55 & 4.0 / 9.5 & 4.0 (0.4--7.6) & 6\% / 31\% \\
Rural & 64 & 5.3 / 16.4 & 7.6 (5.3--9.9) & 12\% / 64\% \\
Urban & 294 & 5.0 / 11.8 & 5.2 (3.2--7.2) & 8\% / 43\% \\
\bottomrule
\multicolumn{5}{p{0.92\textwidth}}{\footnotesize Adjusted difference: silent minus agreement seconds, adjusted for speed, planned lateral acceleration, and vehicle model; route-clustered 95\% CI. Occluded markings (fewer than 30 silent seconds) and wet surface (none) are omitted.}
\end{tabular}
\end{table}

The driving conditions change how often this silence occurs, but not how much the lateral position varies afterwards. Of the not-in-control silent seconds, 59\% occur on urban roads and 11\% on highways, and low-contrast markings account for a larger share of them (31\%) than of all driving time (18\%). However, the extra SDLP after a silent second is between 4 and 8~cm in every marking, lighting, and road-type class, and the confidence intervals of these classes overlap. This means that silence with no one in control happens mostly on urban roads at moderate speed; on highways it is rare, but when it does happen, the lateral position varies just as much.

\subsection{Sensitivity Analysis}
Four sensitivity analyses check whether the main results depend on the analysis choices and on the key assumption of the latent-class model. Table~\ref{tab:sens} reports the status sensitivity under each variation; each row changes one choice of the main specification.

The first analysis changes the threshold that defines low perception confidence. The latent-class model is refit with the openpilot lane-line confidence cut at 0.3, 0.4, 0.6, and 0.7 instead of 0.5. The estimated status sensitivity changes by at most 0.05 for all systems, Toyota, and Hyundai, and stays between 0.71 and 0.86 for Honda. The changes also move in a consistent direction: a stricter cut of 0.3 keeps only the more severely degraded seconds and raises the sensitivity slightly, while a looser cut of 0.7 lowers it. Honda's status remains the most sensitive at every threshold, so the choice of 0.5 does not change the conclusions.

The second analysis changes how the seconds are sampled. Because neighboring seconds are strongly correlated, the data are reduced to every fifth second and the model is refit. The estimates barely change: the sensitivity is 0.49 against 0.47 across all systems, and within 0.02 for Toyota and Hyundai. Honda's estimate (0.95) is less stable, because its reduced sample contains few low-confidence seconds.

The third analysis changes the length of the window for the driver control response, looking 2, 3, and 5~s ahead. The status sensitivity stays between 0.46 and 0.47 across all systems, and changes by no more than 0.02 for each make. The driver-response measures change more: the driver-response risk ratio is 1.72, 1.65, and 1.56, and the ratio between silent and agreement seconds is 4.8, 4.3, and 3.7. The window length therefore changes the size of the driver-response measures, but none of the conclusions.

The fourth analysis relaxes the assumption that the displayed status and the perception confidence make their errors independently of each other. Because both signals look at the same road, their errors may be correlated: a scene that is hard for one camera may also be hard for the other. If so, the two signals agree more often, and a model that assumes independence reads this agreement as accuracy, which makes the status look more sensitive than it is. With only three signals, the strength of this correlation cannot be estimated from the data, so the model is refit with the correlation fixed at 25\%, 50\%, and 100\% of its largest possible value. The estimated status sensitivity falls by no more than 0.12 for all systems, Toyota, and Hyundai, and by 0.21 for Honda, and Honda's status remains the most sensitive. In other words, if the two signals do tend to be wrong together, the status is even less sensitive than the main estimate shows, so the conclusion that it misses about half of the degraded seconds holds or becomes stronger.

\begin{table}[!htbp]
\caption{Sensitivity of the displayed status under alternative analysis choices, by make}\label{tab:sens}
\centering
\small
\setlength{\tabcolsep}{4pt}
\begin{tabular}{llcccc}
\toprule
Analysis & Specification & All systems & Toyota & Hyundai & Honda \\
\midrule
Main specification & Threshold 0.5, every second, & 0.47 & 0.46 & 0.44 & 0.79 \\
 & 3~s window, independent errors & & & & \\
\addlinespace
Perception confidence & 0.3 & 0.50 & 0.48 & 0.48 & 0.86 \\
threshold & 0.4 & 0.49 & 0.48 & 0.46 & 0.81 \\
 & 0.6 & 0.46 & 0.47 & 0.41 & 0.78 \\
 & 0.7 & 0.45 & 0.46 & 0.39 & 0.71 \\
\addlinespace
Sampling & Every fifth second & 0.49 & 0.46 & 0.46 & 0.95 \\
\addlinespace
Response window & 2~s & 0.47 & 0.47 & 0.44 & 0.77 \\
 & 5~s & 0.46 & 0.45 & 0.43 & 0.80 \\
\addlinespace
Error correlation between & 25\% of maximum & 0.42 & 0.39 & 0.37 & 0.72 \\
status and perception & 50\% of maximum & 0.39 & 0.43 & 0.32 & 0.58 \\
confidence & 100\% of maximum & 0.45 & 0.43 & 0.41 & 0.75 \\
\bottomrule
\end{tabular}
\end{table}

\subsection{Implications}
These findings have three implications. First, CANDID-LKA can be used to evaluate the trustworthiness of a proprietary driver-facing status, using only the displayed status, the lane perception confidence of a windshield-camera system, and the driver's steering. Beyond measuring how often the status stays silent, it identifies the silent seconds in which the driver is not in control and measures what follows them. The same approach could support consumer ratings, regulatory testing, and audits after deployment, as well as monitoring of the status while driving. Second, assessments of how well pavement markings can be perceived should measure both how often the status switches to \emph{not detected} and how long it stays there, because worn markings may lengthen outages without making them more frequent, as the Hyundai results suggest. Third, incomplete reporting has a measurable cost: when the driver is not in control during a silent second, the lateral position varies about 60\% more than when the status and the perception confidence agree. We therefore suggest that manufacturers and regulators evaluate not only whether the lane keeping function performs well, but also whether its displayed status reports the function's limitations completely, starting with the silent seconds in which the driver is not in control.

\section{Conclusions}\label{sec:conclusions}
This study asked whether the lane-detection status displayed to the driver can be trusted. Using 39{,}800 seconds of naturalistic driving from 13 vehicle models made by three manufacturers, we evaluated the status with two tests. The internal test examined how often the status reports \emph{not detected} under different driving conditions, and how it switches to \emph{not detected} and back. The external test compared the status with the lane perception confidence of a separately mounted windshield camera and with the driver's steering, using a latent-class model and three measures computed directly from the signals. It then examined the outcomes of the silent seconds, in which the status reads \emph{detected} while the perception confidence is low. These outcomes were compared separately for the seconds in which the driver was in control and the seconds in which the driver was not.

The results show that the displayed status is incomplete. It reports only 47\% of the degraded seconds as \emph{not detected}, compared with 80\% for the third-party low perception confidence, and in half of the low-confidence episodes it never reports \emph{not detected} at all. The incompleteness differs widely across makes: the status sensitivity is 0.44 to 0.46 for Hyundai and Toyota and 0.79 for Honda, and the makes also differ in when their status switches to \emph{not detected} and how long it stays there. In 78\% of the silent seconds, the driver is already in control, mostly steering through a sharp turn. In the other 22\%, the driver is not in control, and the lane keeping is left to the system without any warning. In these seconds, the 3~s SDLP is about 60\% larger than in the agreement seconds in which the driver is also not in control, and this difference appears in every marking, lighting, and road-type condition. In short, a \emph{detected} reading does not guarantee that lane perception is reliable, and the silent seconds in which the driver is not in control are where this matters most.

Several limitations apply to these findings. The latent-class model assumes that the three signals make their errors independently. However, because the production lane keeping function and the third-party vision system observe the same road under the same conditions, their errors may be correlated. The sensitivity analysis shows that such a correlation would make the status look more sensitive than it is, but the main conclusions still hold, and if anything they are conservative. The SDLP and the lateral excursion are computed from the vision system's lane-line readings, which are noisier when the system's confidence is low, as it is during silent seconds. The results are therefore interpreted as an association between silence and lateral movement, relative to the agreement seconds. Even so, the difference is as large where the readings are most reliable, on clear markings and in daylight. The driver control response is also inferred from the steering torque alone, so without external observations it cannot be linked to the real-world reason for each steering input. Finally, because vehicles, routes, regions, and sample sizes differ across makes, the comparisons describe each system in its own operating context and do not isolate causal brand effects, and the one-minute clip length cuts off some long outages.

Future work can build on this study in several directions. One direction is to validate the latent state against scenes labeled by human reviewers, which would measure the accuracy of the displayed status directly. Another is to study the outcomes of silent seconds with an independent measure of the vehicle's path, such as high-precision positioning matched to a lane-level map, to confirm the extra lateral movement and identify the maneuver behind each steering input. A third is to extend the analysis to larger fleets, more regions, and longer recordings, so that the makes can be compared more fairly and long outages can be captured. Finally, the framework can be applied to other driver-facing reports, such as the availability indicators of adaptive cruise control and automatic emergency braking.

\section{Data Availability Statement}
The OpenLKA dataset used in this study is publicly available \cite{wang2025openlka2}. The analysis code and the derived second-level data are available from the corresponding author upon reasonable request.

\bibliography{refs}

\end{document}